\documentclass{article}

\usepackage{arxiv}

\usepackage[utf8]{inputenc} 
\usepackage[T1]{fontenc}    
\usepackage{hyperref}       
\usepackage{url}            
\usepackage{booktabs}       
\usepackage{amsfonts}       
\usepackage{nicefrac}       
\usepackage{microtype}      
\usepackage{lipsum}
\usepackage{graphicx}
\graphicspath{ {./images/} }

\title{Deep learning-based fault identification in condition monitoring}

\author{
 Hariom Dhungana \\
  Western Norway University of Applied Sciences\\
  Bergen, Norway\\
   \texttt{hdhu@hvl.no} \\
   \And
 Suresh Kumar Mukhiya \\
  Western Norway University of Applied Sciences\\
  Bergen, Norway\\
   \texttt{skmu@hvl.no} \\
  \And
Pragya Dhungana \\
  Nepal Telecommunications Authority\\
  Kathmandu, Nepal\\
   \texttt{pdhungana@nta.gov.np} \\
   \And
   Benjamin Karic \\
  Institute for Geoinformatics, University of Münster\\
  Münster, Germany\\
   \texttt{b.karic@uni-muenster.de} \\
}

\begin{document}
\maketitle
\begin{abstract}
Vibration-based condition monitoring techniques are commonly used to identify faults in rolling element bearings. Accuracy and speed of fault detection procedures are critical performance measures in condition monitoring. Delay is especially important in remote condition monitoring and time-sensitive industrial applications. While most existing methods focus on accuracy, little attention has been given to the inference time in the fault identification process. In this paper, we address this gap by presenting a Convolutional Neural Network (CNN) based approach for real-time fault identification in rolling element bearings. We encode raw vibration signals into two-dimensional images using various encoding methods and use these with a CNN to classify several categories of bearing fault types and sizes. We analyse the interplay between fault identification accuracy and processing time. For training and evaluation we use a bearing failure CWRU dataset.
\end{abstract}


\section{Introduction}
In modern industry, rotating equipment plays a crucial role in the production process. Failures with this equipment can lead to substantial economic losses and pose potential risks to operators. Bearing failures alone account for more than 50\% of mechanical defects \cite{chen2023transfer}. Therefore, the development of more effective and intelligent techniques for monitoring bearing health is increasingly vital to ensure machines operate properly and reliably \cite{singh2024dev}. To diagnose internal failures while a machine is operational, analysing relevant external information is essential. Within the field of measurement science, the raw vibration signal stands out as the most valuable and fundamental tool for diagnosing faults in rolling bearings \cite{Aburakhia2022hybrid}, \cite{han2021new}.    

The typical stages of condition monitoring include four submodules: fault detection, fault identification, fault quantification, and fault prognostics \cite{dhungana2024rule} . Once a fault is detected, it is necessary to identify the fault type, including which parts or locations of the bearing indicate degradation during the condition monitoring process \cite{Dhungana2024MachineSystem}.  Fault identification is crucial for effective decision making in condition management, demanding both high accuracy and clear interpretability  \cite{dhungana2024CaseSystem}, \cite{Singh2023InitialSystem}. 

Deep Learning (DL) has revolutionized numerous applications by leveraging large datasets, including computer vision, image analysis and classification, natural language processing, speech recognition, language translation, human activity recognition, social network recommendations, and web search engines \cite{menghani2023efficient} . The use of DL has been continuously applied in predictive maintenance, particularly in the fault identification process to identify patterns and anomalies in vibration data that indicate potential failure component degradation. This allows organizations to schedule maintenance proactively, minimizing downtime and reducing the cost of unexpected breakdowns and enhance overall operational efficiency. The CNNs , initially designed for image classification, are also used in fault identification due to their strong adaptive feature extraction capabilities, making them well-suited for detecting various types of bearing faults \cite{han2021new}, \cite{apeir2024predict}.

In critical fields such as healthcare, aerospace, and industrial automation, where timely decisions can prevent accidents, save lives, and reduce downtime, instant information with slightly less accuracy is more valuable than time-delayed information with higher accuracy \cite{zhang2020enhanced} . Achieving high accuracy with low delay in fault identification allows sufficient time for maintenance decision-making, thereby improving system reliability \cite{Singh2023InitialSystem}. Most of related works focus on accuracy, with little attention to computational complexity and time \cite{zhang2020enhanced}. System delay concept was introduced to evaluate computation time in \cite{Aburakhia2022hybrid}. However, their module used a very sophisticated feature named entropy-based wavelet feature and they used random forest classifier for fault identification.   

In this work we want to avoid intervention of domain knowledge for feature extraction by using DL method. Therefore, we investigate a CNN based various image encodings technique for quick and accurate fault identification  task for quick maintenance decision-making.

\section{Methodology}
Machine learning methods rely on data engineers for feature extraction, while DL performs feature engineering automatically, reducing manual effort and the need for domain-specific knowledge. Here, a CNN model, comprising convolution, pooling, and fully connected layers, is used for fault identification. Convolution is a specialized linear operation used for feature extraction, applying kernels (arrays of numbers called tensors) to input data. The kernel scans the input by performing convolution operations to produce an output tensor, known as a feature map. The pooling layer, typically following a convolution layer, performs a down-sampling operation to reduce the spatial dimensions of feature maps. Fully connected layers, or dense layers, are the final layers of the network. Here, each neuron in one layer is connected to every neuron in the next, making the final classifications based on the features extracted by the convolution layers. The number of fully connected layers varies with the classification depth. 

\subsection{Data encoding:}
Transforming a time series vibration signal into a two-dimensional image can be achieved through various techniques such as spectrograms, wavelets, recurrence plots  \cite{jiang2019recurrence}, Gramian Angular Field (GAF) \cite{wang2015encoding}, Markov Transition Field (MTF) \cite{wang2015imaging}, among others.  
\textbf{Pixel strength:} Converting time series vibration data into pixel strength images involves normalizing the time series, discretizing values into finite states, and mapping these states onto a two-dimensional grid \cite{wen2017newcnn}.
\textbf{GAF:} It converts time series into images using polar coordinate-based matrices, preserving the correlation between the signal and time series \cite{wang2015encoding}. Each matrix element represents a trigonometric function of the angle, with signal segments of length n transformed into n×n images.  
\textbf{MTF:} It encodes Markov transition probabilities sequentially to capture dynamic transition statistics and preserve temporal information. The MTF image size depends on the input signal length and fuzzy kernel size, reflecting the dynamic nature of vibration data. However, excessively large MTF inputs or fuzzy kernels can lead to repetitive data and information loss, making fault detection less effective \cite{wang2015imaging}.
\textbf{Recurrence Plot:}  Recurrence quantification analysis statistically describes small-scale structures in recurrence plots, extracting key features like Recurrence Ratio, Determinism, Laminarity, Maximum Diagonal Length, and entropy \cite{jiang2019recurrence}.
\textbf{GAF-MTF:} In this method both GAF images and MTF images are extracted to capture more complete fault features. The GAF technique preserves more comprehensive fault characteristics, whereas the MTF technique captures fewer fault characteristics but includes more dynamic features. The detail explanation can be found on \cite{han2021new} 

\subsection{CNN architecture and evaluation metrics}
To manage the workload, an optimal CNN was determined using a set of standard parameters such as number of convolution layers, activation functions, epochs, batch etc. After tuning the hyper-parameter of model we found 3 convolution layers, SoftMax as activation function, 150 number of epochs and batch size of 64 gives the optimum model for image classification. The CNN model has a total of 3,314,532 trainable parameters and the network architecture is illustrated in Figure 1. The Flatten layer reshapes the output from these convolution layers into a vector, which then passes through fully connected (Dense) layers. These Dense layers progressively learn to classify images into distinct categories at the output layer. During training, the CNN model is optimized using an Adam optimizer and trained on labelled image data. 

\begin{figure}
\includegraphics[width=\textwidth]{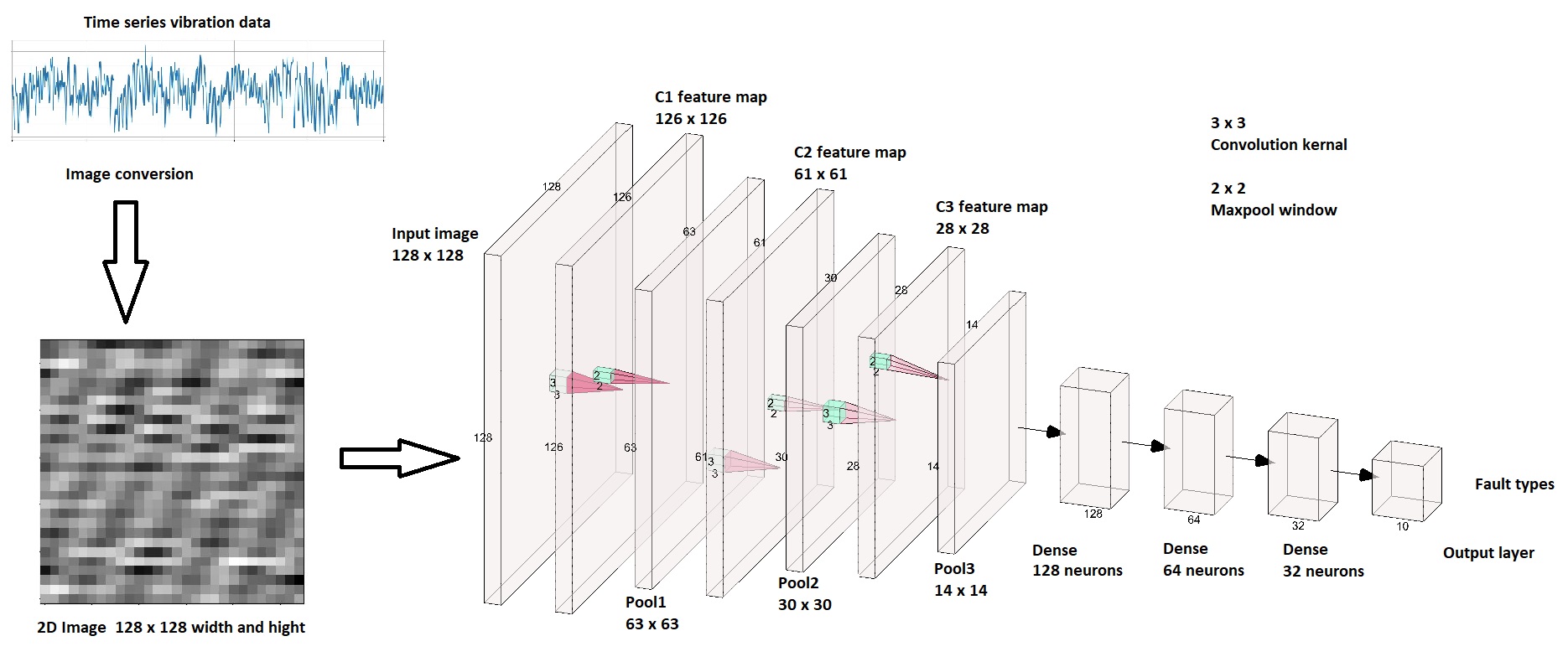}
\caption{Architecture of the proposed CNN for fault identification. Visualization uses AlexNet style.} \label{CNN_structure}
\end{figure}

The performance of fault classification is assessed using two main criteria computation time and accuracy. The confusion matrix, which provides key metrics like accuracy rate, precision, recall, and F-score \cite{petro2024dkdl}.

\subsection{Experimental data sets}
In Case Western Reserve University (CWRU) dataset, experiments were conducted using a 2-hp Reliance Electric motor with vibration data collected via accelerometers \cite{smith2015rolling} . Faults ranging from 0.007 to 0.021 inches in diameter were systematically introduced into the inner raceway (IR), ball, and outer raceway (OR) of bearings. These faulty bearings were then reinstalled in the test motor, and vibration data were recorded at motor loads ranging from 0 to 3 hp (motor speeds from 1720 to 1797 rpm), at a sampling rate of 12,000 samples per second. The drive end acclerometer data are used in this study. Table \ref{tabdataset} details the operational conditions, fault diameters, and motor speeds (rpm) associated with the vibration data.  

We split the vibration measurement data into segments of 1000 data points. A single segment corresponds to 0.0833 seconds of data reading,covering at least two complete bearing rotations. In pixel-based image conversion, the length of the time-domain raw signal \(M^2\) needs to be sufficient to fill an M×M size image. Therefore, 1,000 measurements are sufficient to create a 31x31 image (total of 961 pixels). For all other image encodings the length of raw signal \(M\) is used to fill an M×M size image. To keep encoding time limited a 256 x 256 image (total of 65536 pixels) based on 256 measurements of a segment was used.

The dataset is prelabeled into ten distinct categories. One category represented a healthy sample and the remaining nine categories represented faulty samples that differed with regard to fault diameter and fault location (see Table \ref{tabdataset}). Every vibration file has more than 120,000 vibration measurements for each of the nine faulty classes and more than 480,000 readings for the healthy class. To balance the classes for model training, we used the same proportion of data for each of the four available motor speeds, with 120 segments per class. Each class's data was randomly split into 80\% for training and 20\% for testing for all model trials. We use each dataset for two types of classification: the first is a 4-category classification based on bearing parts, and the second is a 10-category classification based on fault diameter.

\begin{table}[ht]
\centering
\caption{Summary of dataset and fault categories.}\label{tabdataset}
\begin{tabular}{|l|l|cccc|c|c|}
\hline
Health Condition & {Fault diameter} & &Motor speed &(rpm) &  & {4 Class label} & {10 Class label} \\

\hline
Healthy & NaN & 1730 & 1750 & 1772 & 1797 & 1 & 1 \\
\hline
Ball & 0.007" & 1730 & 1750 & 1772 & 1797 & 2 & 2 \\
 & 0.014" & 1730 & 1750 & 1772 & 1797 & 2 & 3 \\
 & 0.021" & 1730 & 1750 & 1772 & 1797 & 2 & 4 \\
\hline
IR & 0.007" & 1730 & 1750 & 1772 & 1797 & 3 & 5 \\
 & 0.014" & 1730 & 1750 & 1772 & 1797 & 3 & 6 \\
 & 0.021" & 1730 & 1750 & 1772 & 1797 & 3 & 7 \\
\hline
OR & 0.007" & 1730 & 1750 & 1772 & 1797 & 4 & 8 \\
 & 0.014" & 1730 & 1750 & 1772 & 1797 & 4 & 9 \\
 & 0.021" & 1730 & 1750 & 1772 & 1797 & 4 & 10 \\
\hline
\end{tabular}
\end{table}


\section{Results and Discussion}
In this section, we evaluate different image encoding techniques. Each type of image has been trained and tested against four different RPM datasets. To compare the training time and accuracy across different conditions, we conducted two sets of experiments. First set experiment contains the four categories (healthy, ball fault, inner race fault and outer race fault). We labelled the three different fault diameters of each component into  a single label (see second last columns in Table \ref{tabdataset}. Second set experiment contains the ten types of faults with unique fault diameter. Four class of fault labels are tested on 4,800 samples and ten class of fault labels are tested on 1,200 samples records. To prevent model bias toward specific fault types they are randomly shuffled and then divided into training and test sets. The test set is used to evaluate the model's performance using one-dimensional pixel strength, GAF, MTF, and recurrence-based images, as well as two-dimensional GAF-MTF images.

\begin{figure}[htb]
\centering
\includegraphics[width=0.8\textwidth]{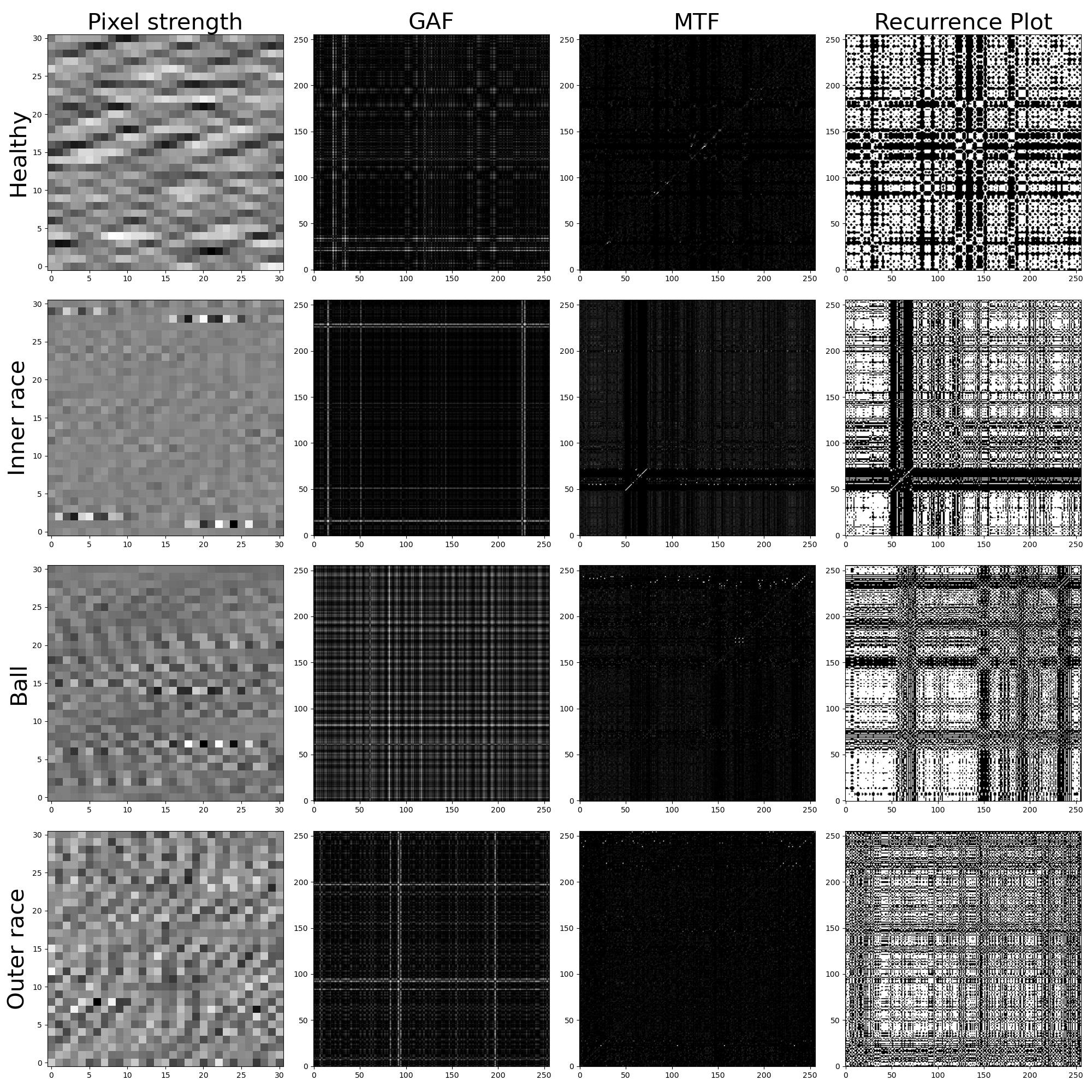}
\caption{Fault images generated by various types of sequence to image methods.} \label{fig:fault_images}
\end{figure}

\subsection{Image conversion}
Figure \ref{fig:fault_images} shows vibration images from four conversion methods and four different bearing conditions with 0.014” faults diameter at 1750 rpm motor speed. Pixel strength based image show distinct patterns corresponding to different fault types. The GAF and recurrence plots display unique vertical and horizontal strips that vary with each type of fault, making it also possible to distinguish between them. However, the MTF exhibits different characteristics, potentially hiding patterns that are difficult to distinguish, except from the case of ball defects. 

\subsection{Classification comparison}
We evaluate image conversion techniques by classification accuracy and total training time, with a table detailing model accuracy and computational time. Confusion matrix offers a detailed breakdown of the model’s classification performance, highlighting true positives, false positives, true negatives, and false negatives. Together, these criteria provide a comprehensive assessment of the model’s effectiveness and efficiency. All analyses are performed on a workstation equipped with an Intel i7 3.0 GHz CPU and 32 GB of RAM. Both image conversion and fault type classification codes are implemented in Python. For the first set of experiments classification (4 categories), accuracy in percentage and training times in seconds are presented in Table \ref{tab3}. Extensive trail under various image conversion methods were conducted, but only the final results from each trail is included in both set of experiments. The same metrics for the second set of experiments (10 categories) are shown in Table \ref{tab2}.

\begin{table}[htb]
\centering
\caption{Total training performance comparison in 4 categories.}\label{tab3}
\begin{tabular}{|l|l|l|l|l|l|l|l|l|l|l|l|}
\hline
Data encoding & {Imag size} & \multicolumn{2}{c|}{1730 rpm} & \multicolumn{2}{c}{1750 rpm} & \multicolumn{2}{c|}{1772 rpm}& \multicolumn{2}{c|}{1797 rpm} & \multicolumn{2}{c|}{All rpm} \\\cline{3-12}
 & & Acc & Time & Acc & Time& Acc & Time& Acc & Time& Acc & Time\\
 & & (\%) & (sec) & (\%) & (sec)& (\%) & (sec)& (\%) & (sec) & (\%) & (sec) \\
\hline
\cline{2-12}
Pixel strength&31 & 98.33 & 23 & 97.91 & 22 & 96.66 & 22 & 96.25 & 23 & 99.16 & 74 \\
\hline
\cline{2-12}
GAF& 256 & 83.75 & 1429 & 72.5 & 1416 & 84.16 & 1272 & 77.08 & 812 & 87.18 & 6835 \\
\hline
\cline{2-12}
MTF& 256 & 68.75 & 1689 & 75 & 1550 & 73.33 & 1535 & 72.08 & 1556 & 80.20 & 4985\\
\hline
\cline{2-12}
Recurrence& 256 & 86.66 & 843 & 81.25 & 848 & 87.08 & 839 & 82.91 & 843 & 93.54 & 3666\\
\hline
\cline{2-12}
GAFMTF& 256 & 87.08 & 1646 & 80 & 1613 & 82.50 & 927 & 77.92 & 1262 & 92.29 & 8068 \\
\hline

\end{tabular}
\end{table}

\begin{table}[htb]
\centering
\caption{Total training performance comparison for 10 category scenario.}\label{tab2}
\begin{tabular}{|l|l|l|l|l|l|l|l|l|l|l|l|}
\hline
Data encoding & {Image size} & \multicolumn{2}{c|}{1730 rpm} & \multicolumn{2}{c}{1750 rpm} & \multicolumn{2}{c|}{1772 rpm}& \multicolumn{2}{c|}{1797 rpm} & \multicolumn{2}{c|}{All rpm} \\\cline{3-12}
 &   & Acc & Time & Acc & Time& Acc & Time& Acc & Time& Acc & Time\\
 &  & (\%) & (sec) & (\%) & (sec)& (\%) & (sec)& (\%) & (sec) & (\%) & (sec) \\
\hline
Pixel strength &31 & 96.25 &23 &97.50 &23 &94.17 &23 &93.33 &23 &96.15 &76 \\
\hline
GAF& 256 & 67.08 & 829 & 64.17 & 832 & 57.08 & 825 & 64.58 & 825 & 72.40 & 3295 \\
\hline
MTF& 256 & 57.52 & 901 & 55.42 & 886 & 55.42 & 886 & 57.92 & 847 & 61.35 & 3565 \\
\hline
Recurrence& 256 & 79.17 & 857 & 80.42 & 858 & 81.25 & 860 & 81.25 & 860 & 86.56 & 3443 \\
\hline
GAFMTF& 256 & 80 & 1656 & 72.5 & 1627 & 67.08 & 1640 & 71.67 & 1636 & 82.6 & 6333 \\
\hline
\end{tabular}
\end{table}

\begin{table}[htb]
\centering
\caption{Inference performance for a single image. The models were trained on data from all motorspeeds and classify 4 fault categories}\label{tab4}
\begin{tabular}{|l|l|l|l|l|l|}
\hline
Data encoding & Image & Data encoding & Model inference & Total inference & Accuracy \\
 & Image size & time (ms) & time (ms) & time (ms) & (\%)  \\
\hline
\cline{2-6}
Pixel strength& 31 & 5.07 & 1.41 & 6.48 & 99.16 \\
\hline
\cline{2-6}
GAF& 256 & 15.99 & 3.02 & 19.01 & 87.18 \\
\hline
\cline{2-6}
MTF& 256 & 37.21 & 3.11 & 40.32 & 80.20 \\
\hline
\cline{2-6}
Recurrence& 256 & 37.15 & 2.96 & 40.11 & 93.54 \\
\hline
\cline{2-6}
GAFMTF& 256 & 56.21 & 3.50 & 59.70 & 92.29 \\
\hline
\end{tabular}
\end{table}

 The CNN models' performance in classifying is then tested, Pixel strength based image conversion is very simple and achieves 96.19\% accuracy in 10-category classification and 99.16 \% accuracy in four-category classification, both in very short computing times of less than 100 seconds. Additionally,  the variation in image size slightly affects computation time. As the image size increases, both computation time and classification accuracy increase proportionally. Moreovere, in both experimental settings of 10-category fault  and 4-category fault, GAF, MTF, and recurrence images require approximately the same computation time, but the accuracy varies. The GAF plot provides better accuracy compared to MTF. GAF transforms the time series data into a polar coordinate system and then into a Gramian matrix, creating images that encode both the magnitude and the relative changes in values. This rich feature representation helps the model learn more informative patterns.  Bearing faults often produce repeating patterns in vibration signals, and recurrence plots are particularly adept at capturing these recurrences, making them highly effective for identifying and distinguishing between different types of bearing faults, therefore, the recurrence plot performs even better than GAF.

By utilizing GAF and MTF images as dual-channel inputs in the neural network, it can be trained to extract both static information from GAF and dynamic information from MTF representations of vibration signals. This combined GAF-MTF image approach yields classification accuracies that are 5–23\% higher than those achieved with MTF alone, with the highest improvement observed in the 1730 rpm trial across both experimental settings. Additionally, the GAF-MTF image combination demonstrates classification accuracies 3–7\% higher than those achieved using GAF alone. The computation time for dual-channel input is slightly higher than that for GAF and MTF, but the accuracy improvement is substantial. 

Combining the four trial datasets of 1730, 1750, 1772, and 1797 rpm makes all rpm trials, which results in 1-3\% higher accuracy during both training and testing compared to training and testing single rpm data separately. This improvement occurs because combining the datasets enhances the diversity of the training data, thereby enabling the model to generalize better to unseen examples during testing. 

From an industry standpoint, the trained model is immediately suitable for use in real-time fault identification, concerning image encoding and inference time. The performance of the presented approach is presented in Table \ref{tab4}. We examined the models that classified four categories of faults and were trained with data from all available motor speeds. The processing time is split up in (1) time for encoding the raw data into an image and (2) performing inference on the created image. Both values are given in milliseconds and combine to the total inference time. The best model in terms of encoding time (5.07 ms), inference time (1.41 ms) and accuracy (99.16\%) is the model using pixel strength as encoding. The model takes only 6.48 ms for fault identification on a data sample of \~80 ms length. Therefore, the proposed pixel strength-based encoding is highly efficient for real-time detection while maintaining high accuracy.

For multi class prediction, the test results can be presented as a 2D confusion matrix to visualize precision, recall, and F1 score. The picture on left side shows the confusion matrix of 10 categories of fault and the picture on right side shows the confusion matrix of 4 categories.  The testing set of each trail consists of 240 images samples from three different fault types (0.007, 0.014, and 0.021) and healthy of 1730 rpm, as shown in Table 1. The ten-category fault classification achieves a 98\% accuracy, while the four-category classification achieves a 97\% accuracy. In the ten-category classification, each fault type is represented with unique damage sizes, resulting in equal proportions for each fault type. In the four-category classification, the damage sizes of 0.007, 0.014, and 0.021 are grouped together into a single fault type. Therefore, the label "2" indicates a healthy condition, while the remaining three labels correspond to faults in the inner race, ball, and outer race, respectively. 

\begin{figure}[htb]
\includegraphics[width=\textwidth]{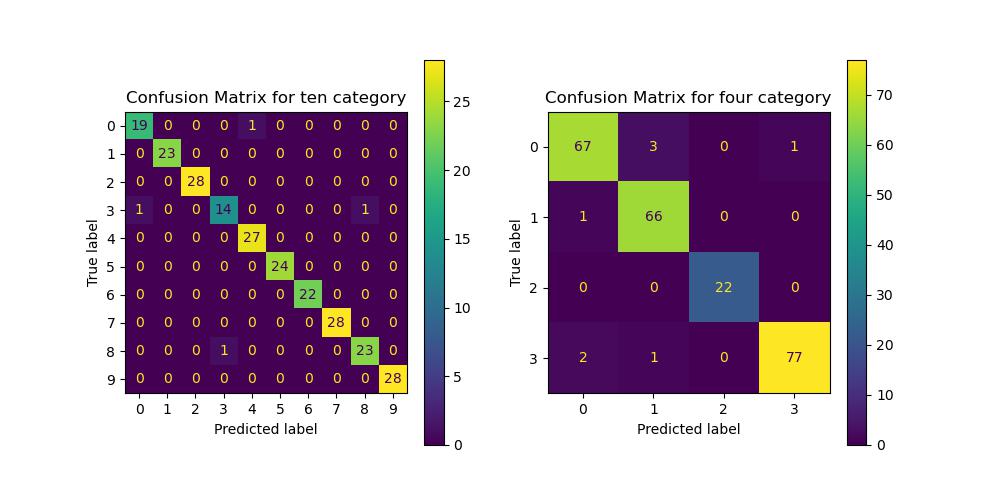}
\caption{A Confusion matrix of fault classification 10 class and 4 classes.} \label{confusion_matrix}
\end{figure}

Recent studies have achieved prediction accuracy's close to 100\% by using complex classification models such as capsule networks and processing long segments of raw signals for image conversion \cite{liang2022intelligent}. However, our goal was to investigate a DL-based initial fault identification model capable of rapidly approximating fault types. Among the five image conversion techniques, the pixel strength-based fault identification model emerges as the most appropriate for quick deployment. 

\section{Conclusion}
The experimental results demonstrate that the standard CNN-based classification method can achieve optimal accuracy in fault identification with minimal computation delay, even under varying loads and motor speeds.  Among those five image conversion techniques, the pixel strength-based image classification model appears suitable for immediate fault identification. Furthermore, incorporating noise reduction techniques in vibration signals or enhancing the CNN classification model with additional layers that demand less computation attention would strengthen this approach for future research. 

\subsubsection{Acknowledgements} 
The authors thank Maneesh Singh, Knut Øvsthus, and Anne-Lena Kampen for their valuable discussions..

\bibliographystyle{unsrt}  


\begin{thebibliography}{1}


\bibitem{chen2023transfer}
Chen, J., Huang, R., Chen, Z., Mao, W. and Li, W., 2023. Transfer learning algorithms for bearing remaining useful life prediction: A comprehensive review from an industrial application perspective. Mechanical Systems and Signal Processing, 193, p.110239.

\bibitem{singh2024dev}
Singh, M., Øvsthus, K., Kampen, A.L. and Dhungana, H., 2024. Development of a human cognition inspired condition management system for equipment. International Journal of System Assurance Engineering and Management, pp.1-10.

\bibitem{Aburakhia2022hybrid}
Aburakhia, S.A., Myers, R. and Shami, A., 2022. A hybrid method for condition monitoring and fault diagnosis of rolling bearings with low system delay. IEEE Transactions on Instrumentation and Measurement, 71, pp.1-13.

\bibitem{han2021new}
Han, B., Zhang, H., Sun, M. and Wu, F., 2021. A new bearing fault diagnosis method based on capsule network and Markov transition field/Gramian angular field. Sensors, 21(22), p.7762.

\bibitem{dhungana2024rule}
Dhungana H., 2024. Rule-Based Decision Making in Biologically Inspired Condition Management System. In Proceedings of the 16th International Conference on Agents and Artificial Intelligence - Volume 3: ICAART; ISBN 978-989-758-680-4, SciTePress, pages 1245-1254. DOI: 10.5220/0012461100003636

\bibitem{Dhungana2024MachineSystem}
Dhungana, P., Singh, R.K. and Dhungana, H., 2023, September. Machine Learning Model for Fault Detection in Safety Critical System. In International Conference on Applications in Electronics Pervading Industry, Environment and Society (pp. 499-507). Cham: Springer Nature Switzerland.

\bibitem{dhungana2024CaseSystem}
Dhungana, H., 2024, April. Case based Decision Making in Biologically Inspired Condition Management System. In 2024 International Conference on Inventive Computation Technologies (ICICT) (pp. 335-339). IEEE.

\bibitem{Singh2023InitialSystem}
Singh, M., Øvsthus, K., Kampen, A.L. and Dhungana, H., 2023, August. Initial Fault Identification for Procedural Decision Making Using Biologically Inspired Condition Management System. In International conference on the Efficiency and Performance Engineering Network (pp. 641-657). Cham: Springer Nature Switzerland.


\bibitem{menghani2023efficient}
Menghani, G., 2023. Efficient deep learning: A survey on making deep learning models smaller, faster, and better. ACM Computing Surveys, 55(12), pp.1-37.

\bibitem{apeir2024predict}
Apeiranthitis, S., Zacharia, P., Chatzopoulos, A. and Papoutsidakis, M., 2024. Predictive Maintenance of Machinery with Rotating Parts Using Convolutional Neural Networks. Electronics, 13(2), p.460.


\bibitem{zhang2020enhanced}
Zhang, Y., Xing, K., Bai, R., Sun, D. and Meng, Z., 2020. An enhanced convolutional neural network for bearing fault diagnosis based on time–frequency image. Measurement, 157, p.107667.

\bibitem{jiang2019recurrence}
Jiang, W., Li, Z., Jiang, A., Lei, Y. and Wang, H., 2019, October. Recurrence plot quantitative analysis-based fault recognition method of rolling bearing. In 2019 Prognostics and System Health Management Conference (PHM-Qingdao) (pp. 1-8). IEEE.

\bibitem{wang2015encoding}
Wang, Z. and Oates, T., 2015, April. Encoding time series as images for visual inspection and classification using tiled convolutional neural networks. In Workshops at the twenty-ninth AAAI conference on artificial intelligence.

\bibitem{wang2015imaging}
Wang, Z. and Oates, T., 2015. Imaging time-series to improve classification and imputation. arXiv preprint arXiv:1506.00327.

\bibitem{wen2017newcnn}
Wen, L., Li, X., Gao, L. and Zhang, Y., 2017. A new convolutional neural network-based data-driven fault diagnosis method. IEEE Transactions on Industrial Electronics, 65(7), pp.5990-5998.


\bibitem{petro2024dkdl}
Petrosyan, O., Pengyi, L., Yulong, H., Jiarui, L., Zhaoruikun, S., Guofeng, F. and Liping, M., 2024. DKDL-Net: A Lightweight Bearing Fault Detection Model via Decoupled Knowledge Distillation and Low-Rank Adaptation Fine-tuning. arXiv preprint arXiv:2406.06653.

\bibitem{smith2015rolling}
Smith, W.A. and Randall, R.B., 2015. Rolling element bearing diagnostics using the Case Western Reserve University data: A benchmark study. Mechanical systems and signal processing, 64, pp.100-131.

\bibitem{liang2022intelligent}
Liang, P., Wang, W., Yuan, X., Liu, S., Zhang, L. and Cheng, Y., 2022. Intelligent fault diagnosis of rolling bearing based on wavelet transform and improved ResNet under noisy labels and environment. Engineering Applications of Artificial Intelligence, 115, p.105269.

\end{thebibliography}

\end{document}